\documentclass[journal]{IEEEtran}
\usepackage{cite}
\usepackage{amsmath,amssymb,amsfonts}
\usepackage{algorithmic}
\usepackage{graphicx}
\usepackage{textcomp}
\usepackage{csquotes}
\usepackage{xcolor,soul}
\usepackage{adjustbox}
\usepackage{hyperref}
\usepackage{multirow}
\usepackage{titlesec}
\usepackage{subcaption}
\usepackage{makecell}
\usepackage{orcidlink}
\usepackage{booktabs} 
\usepackage{caption} 
\usepackage[numbers]{natbib}

\usepackage{etoolbox}
\usepackage{adjustbox}

\definecolor{columbiablue}{rgb}{0.61, 0.87, 1.0}
\titlespacing{\subsubsection}{0pt}{0pt}{1pt}

\begin{document}

\title{Inertial Navigation Meets Deep Learning: A Survey of Current Trends and Future Directions}

\author{Nadav~Cohen \orcidlink{0000-0002-8249-0239}
        and~Itzik~Klein \orcidlink{0000-0001-7846-0654}
\thanks{N. Cohen and I. Klein are with the Hatter Department of Marine Technologies, Charney School of Marine Sciences, University of Haifa, Israel.\\ Corresponding author: ncohe140@campus.haifa.ac.il (N.Cohen)}
}


\maketitle

\begin{abstract}
Inertial sensing is used in many applications and platforms, ranging from day-to-day devices such as smartphones to very complex ones such as autonomous vehicles. In recent years, the development of machine learning and deep learning techniques has increased significantly in the field of inertial sensing and sensor fusion. This is due to the development of efficient computing hardware and the accessibility of publicly available sensor data. These data-driven approaches mainly aim to empower model-based inertial sensing algorithms. To encourage further research in integrating deep learning with inertial navigation and fusion and to leverage their capabilities, this paper provides an in-depth review of deep learning methods for inertial sensing and sensor fusion. We discuss learning methods for calibration and denoising as well as approaches for improving pure inertial navigation and sensor fusion. The latter is done by learning some of the fusion filter parameters.  The reviewed approaches are classified by the environment in which the vehicles operate: land, air, and sea. In addition, we analyze trends and future directions in deep learning-based navigation and provide statistical data on commonly used approaches.
\end{abstract}

\begin{IEEEkeywords}
Inertial sensing, Navigation, Deep learning, Data-driven, Sensor fusion, Autonomous platforms
\end{IEEEkeywords}

\IEEEpeerreviewmaketitle

\section{Introduction}
\begin{figure*}[h] 
  \begin{subfigure}[b]{0.5\linewidth}
    \centering
    \includegraphics[width=0.95\linewidth]{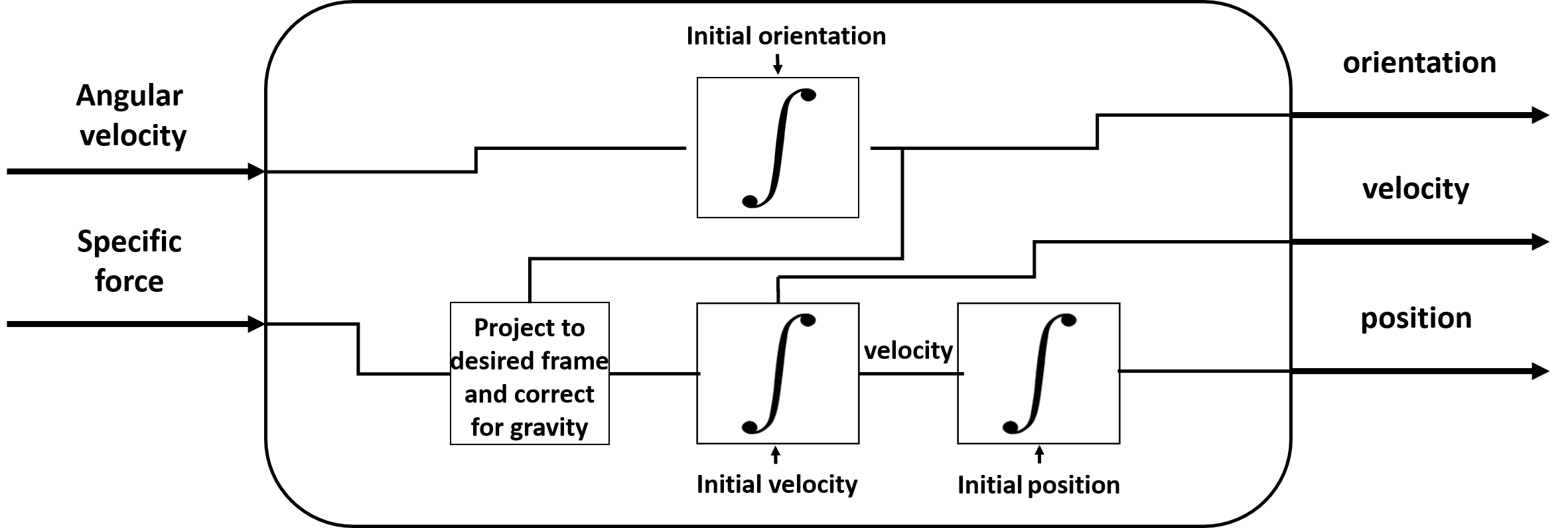} 
        \caption{} 
    \label{fig1:a} 
  \end{subfigure}
  \begin{subfigure}[b]{0.5\linewidth}
    \centering
    \includegraphics[width=0.95\linewidth]{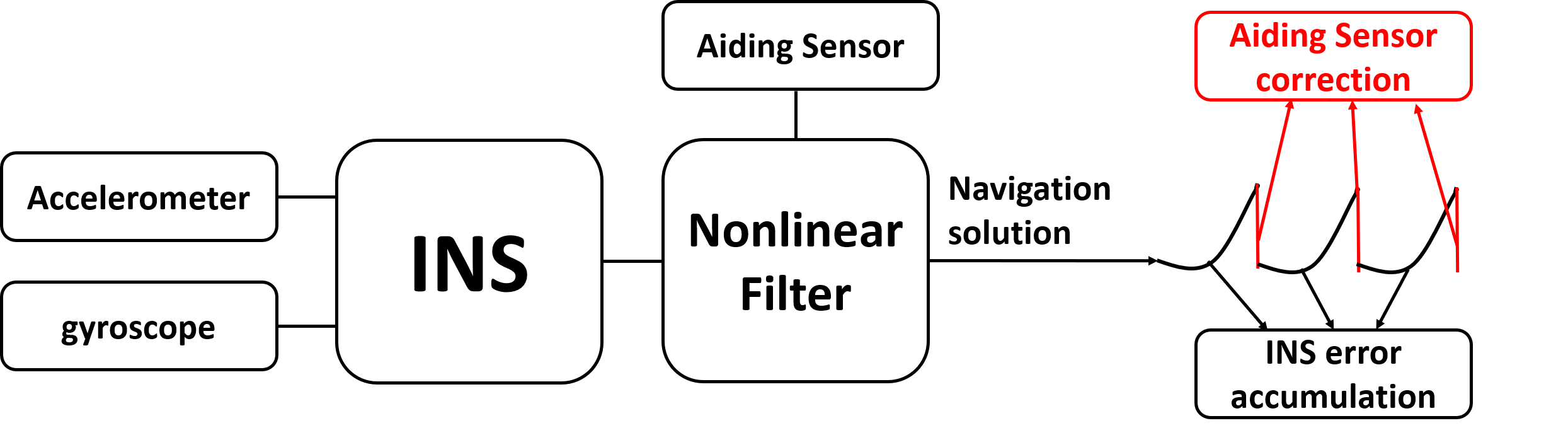} 
        \caption{} 
    \label{fig1:b} 
  \end{subfigure} 
  \caption{(a) Strapdown inertial navigation algorithm. For a given initial condition, the gyroscope's angular velocity and the accelerometer's specific force measurements are integrated over time to calculate the navigation solution in the desired reference frame (local, geographic, and so on). (b) The process by which the navigation solution can be corrected using a nonlinear filter and an aiding sensor. By looking at the output, the black diverging curve shows how the navigation solution accumulates error and the red curve illustrates how the aiding sensor corrects this error, resulting in a chainsaw-like signal.}
\end{figure*}

\IEEEPARstart{R}{esearch} on the concepts of inertial sensing has been conducted for several decades and has been used in the navigation process for a variety of platforms during the last century \cite{mackenzie1993inventing}. As of today, most inertial sensing relies on accelerometers, which provide specific force measurements, as well as gyroscopes, which provide angular velocity measurements \cite{titterton2004strapdown}. 
An inertial measurement unit (IMU) typically consists of three orthogonal accelerometers and three orthogonal gyroscopes, each varying in performance and cost \cite{noureldin2012fundamentals,el2020inertial}. 
\\ \noindent
The IMU readings are processed in real-time to provide a navigation solution. Such a system, which executes the strapdown inertial navigation algorithm, is known as an inertial navigation system (INS) \cite{britting2010inertial}. The INS provides the navigation solution consisting of position, velocity, and orientation as illustrated in Fig.\ref{fig1:a}.
\\ \noindent
The navigation solution accuracy and efficiency is affected by the duration of the mission, the platform dynamics, as well as the quality of the IMU. Using high-end sensors results in a more accurate navigation solution for extended periods of time, while using low-cost INS results in a much faster accumulation of errors. There is, however, a common approach for dealing with error accumulation in both cases, which is to use an accurate external aiding sensor or information aiding to ensure that the solution is bounded or the error is mitigated \cite{farrell2008aided,engelsman2023information}. Sensors like the global navigation satellite system (GNSS), which provides precise position measurements, and the Doppler velocity log (DVL), which offers accurate velocity readings, are considered external aiding sensors. Additionally, supplementary information such as zero velocity updates (ZUPT) and zero angular rates (ZAR) can be utilized, either independently or in conjunction with physical sensors, to mitigate error accumulation in inertial navigation solutions. The INS and aiding sensors are commonly fused in nonlinear filters such as the extended Kalman filter (EKF) and the uncentered Kalman filter (UKF). Filters of this kind have the capability to incorporate the uncertainty of a model, considering both process and measurement noise covariances. This enables them to furnish valuable additional information while also averting error accumulation within an inertial navigation algorithm \cite{groves2015principles}. A diagram of the sensor fusion is shown in Fig. \ref{fig1:b}.
\\ \noindent
During the past decade, deep learning (DL) has made remarkable progress due to advances in neural network architectures, large datasets, and innovative training methods. In the field of computer vision, convolutional neural networks (CNNs) have revolutionized tasks such as image classification, object detection, and semantic segmentation \cite{zaidi2022survey}. It has been demonstrated that recurrent neural networks (RNNs), including long short-term memory (LSTM) networks, are particularly effective at modeling sequences and undertaking language-related tasks, such as translation and sentiment analysis when used for natural language processing \cite{otter2020survey}. Moreover, techniques such as generative adversarial networks (GANs) and transfer learning have extended the capabilities of DL models to enable tasks such as image synthesis and the use of pre-trained models \cite{durgadevi2021generative,zhuang2020comprehensive}. In light of these advancements, DL has been significantly enhanced, propelling it to new frontiers in the field of artificial intelligence.
\\ \noindent
The recent advances in hardware and computational efficiency have proven DL methods to be useful for dealing with real-time applications ranging from image processing and signal processing to natural language processing by utilizing its capabilities to address nonlinear problems \cite{lecun2015deep,goodfellow2016deep,shinde2018review,mahrishi2020machine}.
As a result, DL methods began to be integrated into inertial navigation algorithms. One of the first papers to use neural networks (NNs) in inertial navigation was written by Chiang \emph{et al.} \cite{chiang2003multisensor} in 2003. A multi-sensor integration was proposed using multi-layer, feed-forward neural networks and a back-propagation learning algorithm to regress the accurate land vehicle position. It demonstrated the effectiveness of the NN in addressing navigation problems.\\
Motivated by their success and impact, researchers published papers proposing utilizing deeper and more sophisticated neural networks. Noureldin \emph{et al.} \cite{noureldin2003neuro} designed a multi-layer perceptron (MLP) network to predict the INS position error during GNSS outages using the INS position component and instantaneous time. This work was continued and modified in \cite{sharaf2005online}, where the authors replaced the MLP network with a radial basis function (RBF) neural network to address the same scenario and successfully reduced the position error. In \cite{el2006utilization}, further improvements were described, utilizing multi-layer feed-forward neural networks to regress the vehicle's position and velocity in an INS/differential-GNSS (DGNSS) integration by employing a low-cost IMU. Further research concerning GNSS outage and INS/GNSS fusion is available in \cite{noureldin2011gps} using input-delayed neural networks that regress velocity and position utilizing fully connected (FC) layers. Additionally, in \cite{chen2013novel}, the development of fully connected neural networks with hidden layers constructed of wavelet base functions was proposed to develop INS/GNSS integration to eliminate the complexities associated with KF by providing a reliable positioning solution when GNSS signals are not available.
In addition to position regression, Chiang \emph{et al.} in \cite{chiang2008constructive,chiang2009artificial,chiang2010intelligent} introduced NNs based on fully connected layers for the enhancement of orientation measurements provided by INS/GNSS fusions when using low-cost MEMS IMUs or when GNSS signals are not available. 
Another concept is to improve a specific block within the KF, and a suggestion in \cite{wang2007improving} was to provide the innovation process in an adaptive KF for an integrated INS/GNSS system using a three-layer, fully connected network. Performance analysis
for the networks above was conducted in \cite{malleswaran2013performance} for INS/GNSS fusion. The researchers mentioned above were among the early adopters of utilizing inertial data within deep neural networks (DNNs) to enhance navigation capabilities.
\\ \noindent
Aside from this paper, there are several other papers that conducted surveys on the general topic of data-driven navigation. Most of them looked at specific platforms or DL methods. In \cite{silvestrini2022deep,song2022deep}, a review of the navigation of spacecraft was made in addition to other concepts such as dynamics and control. Furthermore, approaches of only deep-reinforcement learning were reviewed for different platforms in \cite{jiang2020brief,zhu2021deepreinforcement,almahamid2022autonomous}. In light of DL's significant advances in the fields of image processing and computer vision, vision-based navigation surveys were conducted for different platforms and in general \cite{ye2020seeing,zeng2020survey,guastella2020learning,zhu2021deep,tang2022perception}. Some papers focused only on machine-learning-based navigation, which is based primarily on determining the features through preprocessing data analysis \cite{azimi2020survey,li2021inertial,roy2021survey}.
In \cite{golroudbari2023recent} there is a discussion of end-to-end DL methods employed in autonomous navigation, including subjects other than navigation such as obstacle detection, scene perception, path planning, and control. A survey was conducted in \cite{chen2023deep} to examine inertial positioning using recent DL methods. It targeted tasks such as pedestrian dead-reckoning and human activity recognition. An overview of all the survey papers is provided in Table \ref{table:0}.
\begin{table}[h]
\centering
\caption{A collection of fifteen surveys describing DL methods applied to different aspects of navigation.}
\begin{adjustbox}{width=\columnwidth}
\begin{tabular}{|c|c|}
\hline
\textbf{Paper} & \textbf{Topic} \\ \hline
\citenum{silvestrini2022deep}& DL methods for spacecraft dynamics, navigation and control\\ \hline
\citenum{song2022deep}& DL methods of relative navigation of a spacecraft\\ \hline
\citenum{jiang2020brief,zhu2021deepreinforcement}& DRL for mobile robot navigation\\ \hline
\citenum{almahamid2022autonomous}& UAV autonomous navigation using RL\\ \hline
\citenum{ye2020seeing} & DL methods for visual indoor navigation\\ \hline
\citenum{zeng2020survey,zhu2021deep}& Visual navigation using RL\\ \hline
\citenum{guastella2020learning}&  \makecell{DL methods of perception and navigation\\ in unstructured environments} \\ \hline
\citenum{tang2022perception}& DL for perception and navigation in autonomous systems\\ \hline
\citenum{azimi2020survey}& Machine learning for maritime vehicle navigation\\ \hline
\citenum{li2021inertial}& Survey of inertial sensing and machine learning\\ \hline
\citenum{roy2021survey} & Machine learning for indoor navigation\\ \hline
\citenum{golroudbari2023recent} & DL applications and methods for autonomous vehicles \\ \hline
\citenum{chen2023deep}& \makecell{DL methods for positioning including\\ pedestrian dead-reckoning and human activity recognition}\\ \hline
\end{tabular}
\end{adjustbox}
\label{table:0}
\end{table}
\\ \noindent
Contrary to all of the above, this paper examines DL methods utilized exclusively in inertial sensing and sensor fusion algorithms and focuses entirely on vehicles regardless of their operating environment. The contributions of this paper are: 
\begin{enumerate}
    \item Provide an in-depth review of DL methods applied to inertial sensing and sensor fusion tasks for land, aerial, and maritime vehicles.
    \item Examine DL methods for calibrating and denoising inertial sensor data suitable for any vehicle and any inertial sensor. 
    \item Provide insights into current trends on the subject and describe the common DL architectures for inertial navigation tasks.
    \item Discussion of potential future directions for the use of DL approaches for improving inertial sensing and sensor fusion algorithms.
\end{enumerate}

\noindent
The rest of the paper is organized as follows: Sections \ref{land}, \ref{aerial} and \ref{maritime} address DL approaches for improving land, aerial, and maritime inertial sensing, respectively, and are further divided into subsections on pure inertial navigation and aided inertial navigation. 
Section \ref{calib} discusses methods for enhancing the calibration and denoising of inertial data using DL techniques. Section \ref{disc} delves into the survey findings and explores the pros and cons of employing DL in inertial navigation, along with future trends. Lastly, Section \ref{conc} presents the conclusions of this survey..
\section{Land Vehicle Inertial Sensing}\label{land}
\noindent
\subsection{Pure Inertial Navigation}
\begin{figure}[h]
	\centering
        \includegraphics[width=\columnwidth]{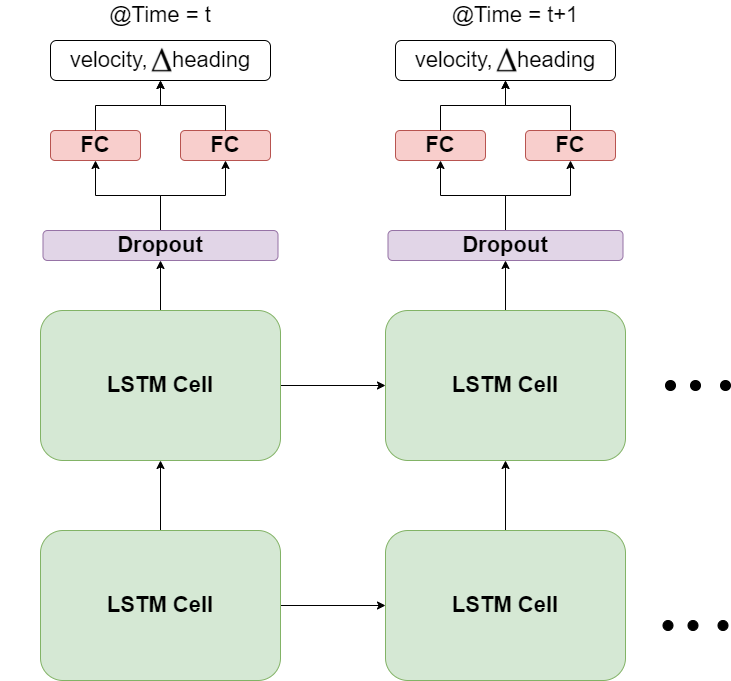}
	  \caption{The figure illustrates an LSTM architecture based on the DeepVIP architecture described in \cite{zhou2022deepvip}. The DeepVIP architecture involves passing inertial readings and additional sensor data through LSTM cells to capture time dependencies. Subsequently, the data traverses dropout layers to prevent overfitting before being processed by FC layers to extract velocity and heading residual outputs.}\label{LSTM}
\end{figure}

\noindent
Several studies have investigated inertial deep learning approaches in situations when GNSS signals are not available. Shen \emph{et al.} in \cite{shen2019dual} presented an approach for improving MEMS-INS/GNSS navigation during GNSS outages. This article proposed two neural networks for a dual optimization process, wherein the first NN compensates for the INS error, while the second NN compensates for the error generated by a filter using the radial basis function (RBF) network for accurate position data.
\\ \noindent
Considering the great interest in scenarios of GNSS outages, many papers have been published suggesting more complex DNNs. In scenarios where GNSS signals are unavailable, Lu \emph{et al.} introduced a multi-task learning method \cite{lu2020heterogeneous}. Initially, inertial data undergo denoising through a convolutional auto-encoder, followed by temporal convolutional network (TCN) processing to address GNSS gaps and one-dimensional CNN (1DCNN) application for zero velocity scenario detection. Subsequently, this aiding data contributes to deriving an accurate navigation solution in Kalman filtering (KF). Additionally, Karlsson \emph{et al.} proposed a CNN model for precise speed estimation solely relying on inertial data in the absence of aiding sensors like GNSS or wheel speed \cite{karlsson2021speed}.
\\ \noindent
GNSS signals are not viable in all scenarios, such as indoor navigation or tunnel navigation, requiring not only compensation for gaps in GNSS signal availability, but also accounting for the entire process. For example, in \cite{tong2021smartphone,tong2022vehicle}, Tong \emph{et al.} regressed the change in velocity and heading of a vehicle in GNSS-blocked environments such as tunnels using a TCN architecture with residual blocks used from low-cost, smartphone mounted, IMU readings. Additionally, "DeepVIP", an LSTM-based architecture, was introduced for indoor vehicle positioning and trained on low-cost inertial data from smartphones. "DeepVIP" is available in two variations. The first achieves a higher level of positioning accuracy by estimating the velocity and change of heading and is appropriate for scenarios requiring the highest degree of positioning accuracy, and the other is more appropriate for situations requiring computational efficiency, therefore its accuracy is slightly lower \cite{zhou2022deepvip}. An illustration based on the DeepVIP model can be seen in Fig. \ref{LSTM}.  \\ \noindent
The use of data-driven methods in inertial vehicle navigation also provides the benefit of improving the output of the inertial sensors independent of the other aiding sensors. Zhao \emph{et al.} examined high-end sensors in \cite{zhao2020learning} and proposed "GyroNet" and "FoGNet", which are based on bidirectional-LSTMs (bi-LSTMs). The first estimates the bias and noise of a gyroscope to improve angular velocity measurements, while the second corrects the drift of the fiber optic gyroscope (FOG) to improve vehicle localization. A similar approach was done in \cite{fei2021research} where the authors introduced a novel approach to produce IMU-like data from GNSS data to train a fully connected-based network to regress angular velocity and acceleration for better positioning based on MEMS IMU data. Gao \emph{et al.} \cite{gao2021glow} introduced the "VeTorch," an inertial tracking system that employs smartphone-derived inertial data for real-time vehicle location tracking. Employing a TCN, they conducted acceleration, orientation sequence learning, and pose estimation. In a separate study, Freydin \& Or proposed an LSTM-based model to forecast vehicle speed using low-cost IMU readings from smartphones \cite{freydin2022learning}.
\\ \noindent
Apart from enhancing inertial reading abilities for more accurate navigation, DL approaches have demonstrated a significant impact in improving sensor fusion. Li \emph{et al.} \cite{li2019deep} introduced a novel recurrent convolutional neural network (RCNN)-based architecture for scan-to-scan laser/inertial data fusion for pose estimation. Additionally, Srinivasan \emph{et al.} \cite{srinivasan2020end} proposed an end-to-end RNN-based approach that utilizes IMU data along with wheel odometry sensor and motor current data to estimate velocity, following an investigation into the extreme dynamics of an autonomous race car. In the "GALNet" framework, the authors utilized inertial, kinematic, and wheel velocity data to train an LSTM-based network that regressed the relative pose of a car \cite{mendoza2020galnet}. In \cite{zhou2022xdrnet}, a smartphone gyroscope, magnetometer, and gravity sensor were integrated to train "XDRNet" architecture. Using 1DCNN, the network regresses vehicle speed and heading changes and, as a consequence, reduces inertial positioning error drift. Moreover, Liu \emph{et al.} presented a hybrid CNN-LSTM-based network that integrates a dual-antenna satellite receiver and MEMS IMU to forecast the residual of the position, velocity, and orientation at each time step \cite{liu2022integrated}. In the case of an agricultural wheeled robot, velocity estimation was achieved by integrating inertial readings, magnetometer data, and the absolute values of the mean discrete Fourier transform coefficients of accelerometer norms within a TCN architecture. These velocity estimates were subsequently utilized in INS/GNSS fusion to estimate 2D position and velocity, reducing the reliance on GNSS measurements, particularly in GNSS-denied environments, and demonstrating good performance in such scenarios \cite{du2023neural}.

\subsection{Aided Inertial Navigation}
\noindent
It has been demonstrated that using a filter such as the extended KF (EKF) is a common approach to achieving a high level of accuracy and reliability in sensor fusion. Aside from providing a navigation solution, the filter also offers insight into the propagation of navigation uncertainty over time. Consequently, DL methods have shown great impact in addressing different aspects of the filter that significantly influence the solution. Hosseinyalamdary proposed a deep KF \cite{hosseinyalamdary2018deep}, which includes a modeling step alongside the prediction and update steps of the EKF. This addition corrects IMU positioning and models IMU errors, with GNSS measurements used to learn IMU error models using RNN and LSTM methods. In the absence of GNSS observations, the trained model predicts the IMU errors. Furthermore, SL-SRCKF (self-learning square-root-cubature KF) employs an LSTM-based network to continuously obtain observation vectors during GNSS outages, learning the relationship between observation vectors and internal filter parameters to enhance the accuracy of integrated MEMS-INS/GNSS navigation systems \cite{shen2020seamless}. Additionally, DL methods identify specific scenarios in the inertial data that may prevent the navigation solution from accumulating errors. Using RNN, a zero velocity situation or no lateral slip could be identified and incorporated later on into a KF for localization processes \cite{brossard2019rins}.
\\ \noindent
The literature indicates that the covariance noise matrix plays an important role in the KF, and therefore DL methods were employed to adapt it consistently. According to Brossard, Barrau, \& Bonnabel, a CNN-based method was used to dynamically adapt the covariance noise matrix for an invariant-EKF using moderate-cost IMU measurements \cite{brossard2020ai}. Previously, in \cite{brossard2019learning}, the authors devised a method that combined Gaussian processes with RBF neural networks and stochastic variational inference. This approach aimed to enhance a state-space dynamical model's propagation and measurement functions by learning residual errors between physical predictions and ground truth. Moreover, the study demonstrated how these corrections could be utilized in the design of EKFs.
An alternative method of estimating the process noise covariance relies on reinforcement learning, as explained in \cite{gao2020rl}, which uses an adaptive KF to determine position, velocity, and orientation. In \cite{wu2020predicting}, not only the parameters of measurement noise covariances but also the parameters of process noise covariances were regressed. These parameters can be more accurately estimated using a multitask TCN, resulting in higher position accuracy than traditional GNSS/INS-integrated navigation systems. The authors in \cite{xiao2021residual} introduced a residual network incorporating an attention mechanism to predict individual velocity elements of the noise covariance matrix. As a result of the empirical study, it has been demonstrated that adjusting the non-holonomic constraint uncertainty during large dynamic vehicle motions rather than strictly setting the lateral and vertical velocities to zero could improve positioning accuracy under large dynamic motions.

\begin{table}[h!]
\centering
\caption{A summary of thirty-five papers describing DL inertial sensing and sensor fusion for land vehicles, categorized by their improvement goals.}
\begin{adjustbox}{width=\columnwidth}
\tiny
\begin{tabular}{|c|c|}
\hline
Improvement Goals& Papers \\
\hline
Position         &\makecell{\citenum{chiang2003multisensor,noureldin2003neuro,sharaf2005online,el2006utilization,chiang2008constructive,chiang2009artificial,chiang2010intelligent,lu2020heterogeneous}\\\citenum{noureldin2011gps,chen2013novel,malleswaran2013performance,shen2019dual,fei2021research,gao2021glow,li2019deep,mendoza2020galnet}\\\citenum{liu2022integrated}}  \\ \hline
Velocity         & \makecell{\citenum{el2006utilization,noureldin2011gps,zhao2020learning,srinivasan2020end,tong2021smartphone,tong2022vehicle,karlsson2021speed,lu2020heterogeneous}\\\citenum{zhou2022xdrnet,zhou2022deepvip,liu2022integrated,freydin2022learning,du2023neural}} \\ \hline
Orientation       & \makecell{\citenum{chiang2008constructive,chiang2009artificial,chiang2010intelligent,zhao2020learning,mendoza2020galnet,tong2021smartphone,tong2022vehicle,lu2020heterogeneous}\\\citenum{fei2021research,gao2021glow,li2019deep,zhou2022deepvip,zhou2022xdrnet,liu2022integrated,brossard2019rins}} \\ \hline
Filter Parameters & \makecell{\citenum{wang2007improving,hosseinyalamdary2018deep,shen2019dual,brossard2019learning,gao2020rl,brossard2020ai,wu2020predicting,xiao2021residual}\\\citenum{shen2020seamless}}  \\ \hline
\end{tabular}
\end{adjustbox}
\end{table}

\section{Aerial Vehicle Inertial Sensing}\label{aerial}
\noindent
\subsection{Pure Inertial Navigation}
\noindent
A number of the papers examined visual-inertial odometry as a tool for aerial inertial navigation. Clark \emph{et al.} \cite{clark2017vinet} developed the "VINet" architecture. It comprises LSTM blocks that process camera output at the camera rate and IMU LSTM blocks that process data at the IMU rate. This study used deep learning to determine a micro air vehicle's (MAV's) orientation. In \cite{baldini2020learning}, a similar approach was employed; however, the platform was not an MAV, but rather a bigger and heavier unmanned aerial vehicle (UAV). Another vision-inertial fusion study was conducted in \cite{aslan2022hvionet} in which a camera and IMU sensor fusion method were used to estimate the position of an unmanned aircraft system (UAS) using a CNN-LSTM-based network known as "HVIOnet", which stands for hybrid visual-inertial odometry network. A more complex method was introduced by Yusefi \emph{et al.} where an end-to-end, multi-model, DL-based, monocular, visual-inertial localization system was utilized to resolve the global pose regression problem for UAVs in indoor environments. Using the proposed deep RCNN, experimental findings demonstrate an impressive degree of time efficiency, as well as a high degree of accuracy in UAV indoor localization \cite{yusefi2021lstm}.

\begin{figure}[h]
	\centering
        \includegraphics[width=\columnwidth]{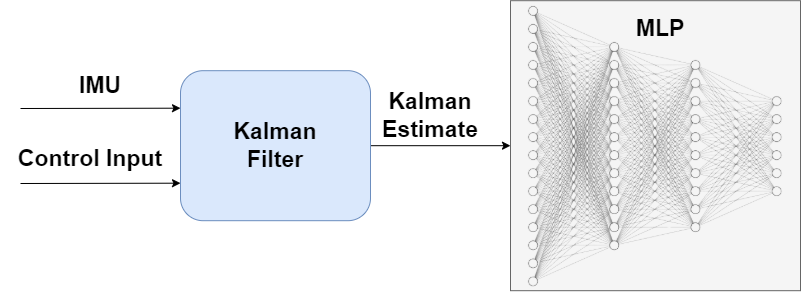}
	  \caption{The figure illustrates the architecture presented in \cite{al2019deep}, which is based on a simple MLP network. This network utilizes the estimated attitude states from the Kalman filter and enhances them through training with data containing accurate reference attitude information.}\label{MLP}
\end{figure}

\noindent
For assessing a vehicle's attitude with only inertial sensing, Liu \emph{et al.} employed an LSTM-based network based on IMU data from a UAV \cite{liu2018attitude}. While in \cite{silva2019end}, the authors utilized a hybrid, more complex network that incorporates CNN and LSTM blocks to estimate MAV pose utilizing the current position and unit quaternion. Esfahani \emph{et al.} introduced an inertial odometry network named "AbolDeepIO" \cite{esfahani2019aboldeepio}. Based on LSTM, this network architecture leverages IMU readings for inertial odometry in MAVs. In their study, they compared the performance of AbolDeepIO with that of "VINet" \cite{clark2017vinet}, demonstrating superior results in MAV data analysis. Seven sub-architectures were also tested and evaluated. As a follow-up to "AbolDeepIO", the authors brought forward "OriNet", which is also based on LSTM and capable of estimating the full 3D orientation of a flying robot with a single particular IMU in the quaternion form \cite{esfahani2019orinet}. A robust inertial attitude estimator called "RIANN" was also proposed, a network whose name stands for Robust IMU-based Attitude Neural Network. Several motions, including MAV motion, were regressed using a gated recurrent units (GRUs) network. As well as proposing three domain-specific advances in neural networks for inertial attitude estimation, they also propose two methods that enable neural networks to handle a wide variety of sampling rates \cite{weber2021riann}. It has been suggested by Chumuang \emph{et al.} \cite{chumuang2022feature} that CNNs and LSTMs are both effective for predicting the orientation of a MAV, the former using the quaternions predicted from data from the IMU using Madgwick's adaptive algorithm \cite{madgwick2011estimation}, and the latter using raw gyroscope measurements. For improving MAV navigation, rather than using two separate networks, the authors in \cite{golroudbari2023end} suggested three models including CNN, RNN, and a CNN and LSTM hybrid model, and performed a comparison amongst them.
\\ \noindent
In aerial navigation, as in other areas, GNSS signals can be useful when integrating with the INS, and in the absence of these signals, the inertial navigation solution drifts. To achieve better performance than traditional GNSS/INS fusion, an LSTM-based network was proposed in \cite{narkhede2022inertial} to estimate the 3D position of an aerial vehicle. When the aerial vehicle encounters GNSS-denied environments, DL approaches are used to compensate. Continuing their research in \cite{liu2018attitude}, as discussed above, Liu \emph{et al.} proposed a 1DCNN/GRU hybrid deep learning model that predicts the GNSS position increments for integrated INS/GNSS navigation in the event of GNSS outages\cite{liu2021gps,liu2021novel,liu2022deep}. A similar approach to deal with scenarios of denied GNSS environments was implemented by \cite{geragersian2022ins}, in which a GRU-based network was used to estimate position and velocity. A novel method known as "QuadNet" was proposed by Shurin \& Klein in \cite{shurin2022quadnet}. They enforced the quadrotor motion to be periodic and utilized its inertial readings to develop 1DCNN and LSTM methods for regression of the distance and altitude changes of the quadrotor. In the study by Hurwitz and Klein, the "QuadNet" architecture was revisited to explore the benefits of using multiple IMUs and to devise effective methods for leveraging the excess inertial data. In a different paper and scenario involving GNSS-denied environments, the researchers employed a combination of optical odometry, radar height estimates, and multi-sensory data fusion. To enhance optical flow velocity estimates in these challenging conditions, they utilized an LSTM network alongside angular velocity readings from the IMU to predict velocity increments \cite{deraz2023deep}.
\subsection{Aided Inertial Navigation}
\noindent
To enhance the quality of Kalman attitude estimates, Al-Sharman \emph{et al.} proposed a fully connected network in \cite{al2019deep}. This network takes the Kalman state estimates, derived from inertial readings and control vectors, as inputs, and regresses the UAV attitude. Figure \ref{MLP} illustrates the approach based on the one presented in the aforementioned paper. Another recurring approach involves predicting noise covariance information using DL. In \cite{zou2020cnn}, the authors introduced a CNN-based adaptive Kalman filter designed to enhance high-speed navigation with low-cost IMUs. Their approach employs a 1DCNN to predict noise covariance information for 3D acceleration and angular velocity, utilizing windowed inertial measurements. The aim is to outperform classical Kalman filters and Sage-Husa adaptive filters in high dynamic conditions. In a subsequent paper, Or \& Klein \cite{or2022hybrid} developed a data-driven, adaptive noise covariance approach for an error state EKF in INS/GNSS fusion. Using a 1DCNN, they were able to estimate the process noise covariance matrix and use the information to provide a better navigation solution for a quadrotor drone. Solodar and Klein \cite{solodar2023vio}  proposed VIO-DualProNet, consisting of two 1DCNNs tasked with estimating the covariance matrices of the accelerometer and gyroscope, respectively. These estimates are utilized to dynamically assess the process noise covariance matrix, enhancing the optimization process in visual-inertial odometry. Experimental findings on MAVs demonstrate a 25\% enhancement in absolute trajectory error compared to the baseline approach.

\begin{table}[h!]
\centering
\caption{A summary of twenty-two papers describing DL inertial sensing and sensor fusion for aerial vehicles, categorized by their improvement goals.}
\begin{adjustbox}{width=\columnwidth}
\begin{tabular}{|c|c|}
\hline
Improvement Goals& Papers \\
\hline
Position         &\makecell{\citenum{aslan2022hvionet,golroudbari2023end,narkhede2022inertial,liu2021gps,liu2021novel,liu2022deep,geragersian2022ins,shurin2022quadnet}\\\citenum{yusefi2021lstm,silva2019end}}  \\ \hline
Velocity         & \citenum{golroudbari2023end,geragersian2022ins} \\ \hline
Orientation       & \makecell{\citenum{clark2017vinet,baldini2020learning,yusefi2021lstm,liu2018attitude,silva2019end,esfahani2019aboldeepio,weber2021riann,esfahani2019orinet}\\\citenum{chumuang2022feature,golroudbari2023end,shurin2022quadnet,deraz2023deep}} \\ \hline
Filter Parameters & \citenum{al2019deep,zou2020cnn,or2022hybrid,solodar2023vio}  \\ \hline
\end{tabular}
\end{adjustbox}
\end{table}

\section{Maritime Vehicle Inertial Sensing}\label{maritime}
\begin{figure*}[h]
	\centering
        \includegraphics[width=\textwidth]{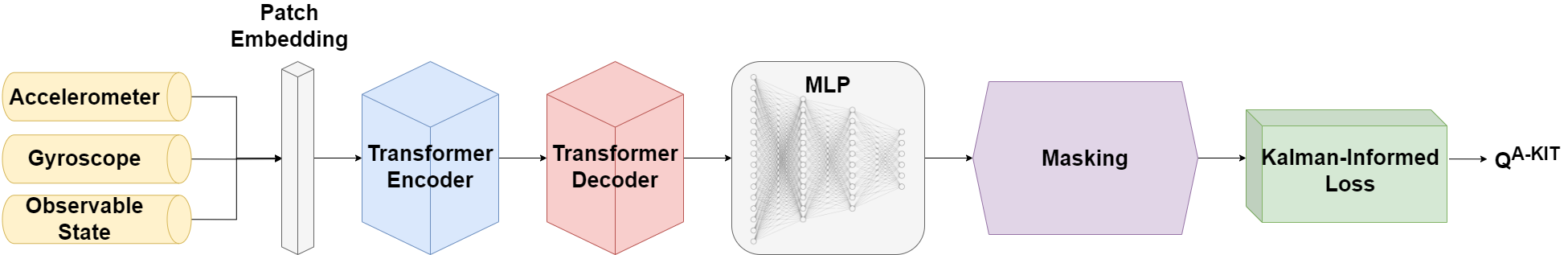}
	  \caption{The figure depicts the architecture based on A-KIT introduced in \cite{cohen2024kit}, where an adaptive Kalman-informed transformer is employed. In this approach, inertial data along with observable states, such as position from GNSS, are fed through the block diagram presented. The output obtained from this process provides the scale factors required for dynamically estimating the process noise covariance matrix of the extended Kalman filter. }\label{A-KIT}
\end{figure*}
\noindent
\subsection{Pure Inertial Navigation}
\noindent
Zhang \emph{et al.} \cite{zhang2019application,mu2019end} were the first to apply deep learning techniques to autonomous underwater vehicle (AUV) navigation. To regress the position displacements of an AUV, the authors used an LSTM-based network trained over GNSS data that was utilized as the position displacement training targets. The output of this network is incorporated into EKF for the purpose of making the position directly observable. Later, the authors developed a system called "NavNet", which utilizes data from the IMU and DVL sensors through a deep learning framework using LSTM and attention mechanisms to regress the position displacement of an AUV and compares it to the EKF and UKF \cite{zhang2020navnet}. Further improvements and revisions were made in \cite{zhang2023sequential}, which used TCN blocks instead of LSTM blocks. As a way of rectifying the error accumulation in the navigation system, Ma \emph{et al.} proposed a similar procedure, as previously mentioned, and developed an adaptive navigation algorithm for AUV navigation that uses deep learning to generate low-frequency position information as a method of generating low-frequency positioning information. Based on LSTM blocks, the network receives velocity measurements from the DVL and Euler angles from the attitude and heading reference system (AHRS) \cite{ma2021adaptive}. Weizman \emph{et al.} \cite{weizman2023enhancement} investigated the maneuvers of ocean gliders and utilized their periodic motion to introduce GilderNet, a 1DCNN that regresses glider distance and depth based solely on inertial sensors and inspired by methods from pedestrian dead reckoning.
\subsection{Aided Inertial Navigation}
\noindent
A task such as navigation on or underwater may encounter difficulties due to the dynamics of the environment or the inaccessibility of GNSS signals. In \cite{he4405121tcn} the authors used a hybrid TCN-LSTM network to predict the pitch and heave movement of a ship in different challenging scenarios. A fully connected network is suggested in \cite{song2020neural} to improve AUV navigation in rapidly changing environments, such as in waves near or on the surface. This network is based on data from an accelerometer and is used to predict the pitch angle. The Kalman filter, neural network, and velocity compensation are then combined in the "NN-DR" method to provide a more accurate navigation solution.
\\ \noindent
The DVL is used in underwater applications as a sensor to assist in navigation, similar to GNSS data used by above-water applications. Several papers have been published examining scenarios of DVL failure. There was, for example, a proposal in \cite{saksvik2021deep} to aid dead-reckoning navigation for AUVs with limited sensor capabilities. Using RNN architecture, the algorithm predicts relative velocities based on the data obtained from the IMU, pressure sensors, and control actions of the actuators on the AUV. In \cite{li2021underwater} the Nonlinear AutoRegressive with Exogenous Input (NARX) approach is used in cases where DVL malfunctions occur to determine the velocity measurements using INS data. For receiving the navigation solution, the output of the network is integrated into a robust Kalman filter (RKF). Cohen \& Klein proposed "BeamsNet", which replaces the model-based approach to derive the AUV velocity measurements out of the DVL raw beam measurements using 1DCNN that uses inertial readings \cite{cohen2022beamsnet}. The authors continued the work by looking at cases of partial DVL measurements and succeeded in recovering the velocity with a similar architecture called "LiBeamsNet" \cite{cohen2022libeamsnet}.
 In the case of a complete DVL outage, in \cite{cohen2022set} the authors introduced "ST-BeamsNet", which is a Set-Transformer based network that uses inertial reading and past DVL measurements to regress the current velocity.\\
 Lately, Topini \emph{et al.} \cite{topini2023experimental} conducted an experimental comparison of data-driven strategies for AUV navigation in DVL-denied environments where they compared MLP, CNN, LSTM, and hybrid CNN-LSTM networks to predict the velocity of the AUV. 
\noindent
 According to the authors of \cite{shaukat2021multi}, the RBF network can be augmented with error state KF to improve the state estimation of an underwater vehicle. Through the application of the RBF neural network, the proposed algorithm compensates for the lack of error state KF performance by enhancing innovation error terms.
 Or \& Klein  \cite{or2022hybrid, or2022adaptive, or2023pronet} developed an adaptive EKF for velocity updates in INS/DVL fusions. Initially, they demonstrated that by correcting the noise covariance matrix using 1DCNN to predict the variance in each sample time using a classification problem, they could significantly improve navigation results. In recent work, the authors introduced "ProNet", which uses regression instead of classification to accomplish the same task. In \cite{cohen2024kit}, an adaptive Kalman-informed transform (A-KIT) was presented. This method utilized inertial readings along with Kalman estimates of the velocity vector within a transformer-based network to regress the scale factors required for adjusting the process noise covariance matrix. The results demonstrated superior performance compared to standard EKF and various adaptive versions. Additionally, a Kalman-informed loss was introduced to ensure that the output aligns with the Kalman theory. The A-KIT approach is illustrated in Fig \ref{A-KIT}. 

\begin{table}[h!]
\centering
\caption{A summary of nineteen papers describing DL inertial sensing and sensor fusion for maritime vehicles, categorized by their improvement goals.}
\begin{adjustbox}{width=\columnwidth}
\tiny
\begin{tabular}{|c|c|}
\hline
Improvement Goals& Papers \\
\hline
Position         &\citenum{zhang2019application,mu2019end,zhang2020navnet,zhang2023sequential,ma2021adaptive,he4405121tcn,song2020neural,weizman2023enhancement}  \\ \hline
Velocity         & \citenum{song2020neural,saksvik2021deep,cohen2022beamsnet,cohen2022libeamsnet,cohen2022set,topini2023experimental,li2021underwater} \\ \hline
Orientation       & \citenum{he4405121tcn,song2020neural} \\ \hline
Filter Parameters & \citenum{shaukat2021multi,or2022hybrid, or2022adaptive, or2023pronet,cohen2024kit}  \\ \hline
\end{tabular}
\end{adjustbox}
\end{table}

\section{Calibration and Denoising}\label{calib}
\begin{figure}[h]
	\centering
        \includegraphics[width=\columnwidth]{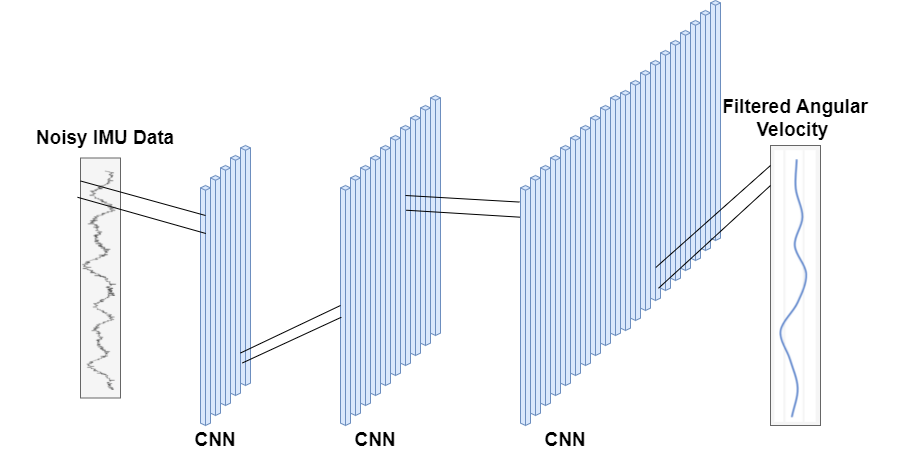}
	  \caption{The figure is based on the architecture introduced in \cite{chen2018improving}, where a CNN architecture is proposed to address IMU errors by analyzing a window of inertial measurements, enabling the detection and removal of noisy features. The block diagram depicts noisy data passing through convolutional layers, with subsequent smoothing or filtering of the input.}\label{CNN}
\end{figure}

\noindent
Since the INS is based on the integration of inertial data over time, it accumulates errors due to structured errors in the sensors. Calibration and denoising are crucial to minimizing these errors. In Chen \emph{et al.} \cite{chen2018improving}, deep learning was used for the first time to reduce IMU errors. IMU data, containing deterministic and random errors, is fed as input to CNN, which filters the data, an illustration van be seen in Fig. \ref{CNN}.
According to Engelsman \& Klein \cite{engelsman2022data,engelsman2022learning}, an LSTM-based network can be used for de-noising accelerometer signals and a CNN-based network can be used to eliminate bias in low-cost gyroscopes.
\\ \noindent
Apart from the papers mentioned above, the majority of research focuses on the denoising and calibration of gyroscopes. A series of papers \cite{jiang2018performance,jiang2018mems,zhu2019mems} examined the denoising of gyroscope data by utilizing various variations of RNNs. One paper demonstrated the performance of a simple RNN structure, while the others utilized LSTMs. A comparison was made between LSTM, GRU, and hybrid LSTM-GRU approaches for gyroscope denoising in \cite{jiang2019mixed}. Additional comparisons between GRU, LSTM, and hybrid GRU-LSTM were conducted in \cite{han2021hybrid}. In Brossard \emph{et al.} \cite{brossard2020denoising}, a deep learning method is presented for reducing the gyroscope noise in order to achieve accurate attitude estimations utilizing a low-cost IMU. For feature extraction, a dilated convolutional network was used and for training on orientation increments, an appropriate loss function was utilized. Various CNN-based architectures have been explored to address gyroscope corrections. One study showcased a denoising autoencoder architecture, constructed on a deep convolutional model, aimed at restoring clean and undistorted output from corrupted data. It was found that the KF angle prediction was boosted in this scenario \cite{russo2020danae}. Furthermore, a TCN and 1DCNN were integrated for MEMS gyroscope calibration in \cite{huang2022mems}. Liu \emph{et al.} introduced "LGC-Net" as a method for extracting local and global characteristics from IMU measurements to regress gyroscope compensation components dynamically. This model utilizes special convolution layers and attention mechanisms for this purpose \cite{liu2022lgc}. Yuan \emph{et al.} proposed "IMUDB" as a self-supervised IMU denoising method inspired by natural language processing techniques. This approach addresses the challenge of obtaining sufficient and accurate annotations for supervised learning while achieving promising results \cite{yuan2023simple}. Engelsman \emph{et al.} developed a bidirectional LSTM-based network specifically for gyrocompassing, particularly suitable for low-performance gyroscopes affected by the limited signal strength of Earth's rotation rate, which is often overshadowed by gyro noise \cite{engelsman2023towards}. They subsequently demonstrated the effectiveness of this approach in unmanned underwater vehicle applications \cite{engelsman2024underwater}.

\section{Discussion}\label{disc}
\begin{figure*}[htbp] 
  \begin{subfigure}[b]{0.5\linewidth}
    \centering
    \includegraphics[width=\columnwidth]{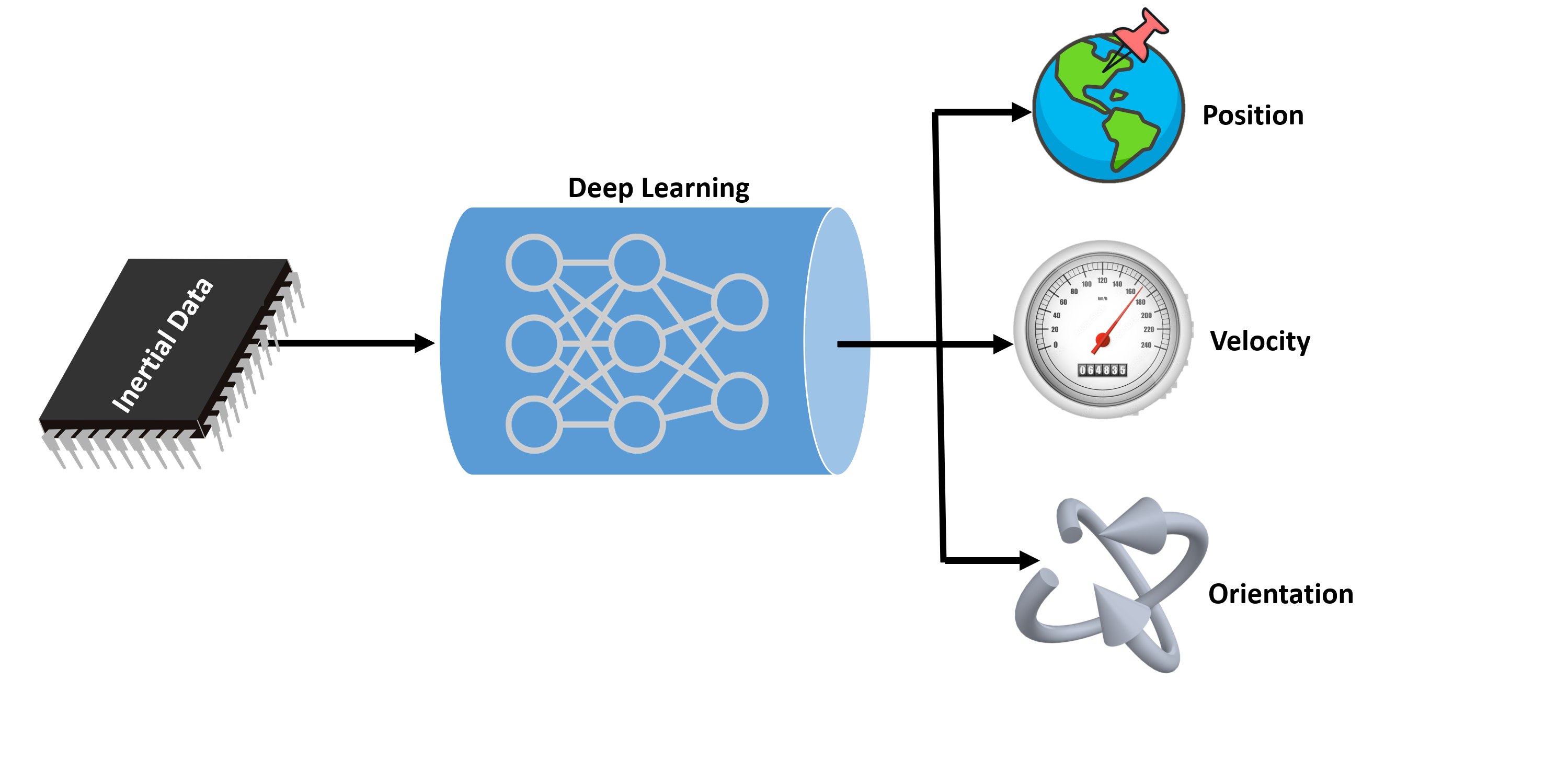} 
    \caption{Using an end-to-end DL approach to regress the full \\ state of the vehicle.} 
    \label{fig2:a} 
    \vspace{2ex}
  \end{subfigure}
  \begin{subfigure}[b]{0.5\linewidth}
    \centering
    \includegraphics[width=\columnwidth]{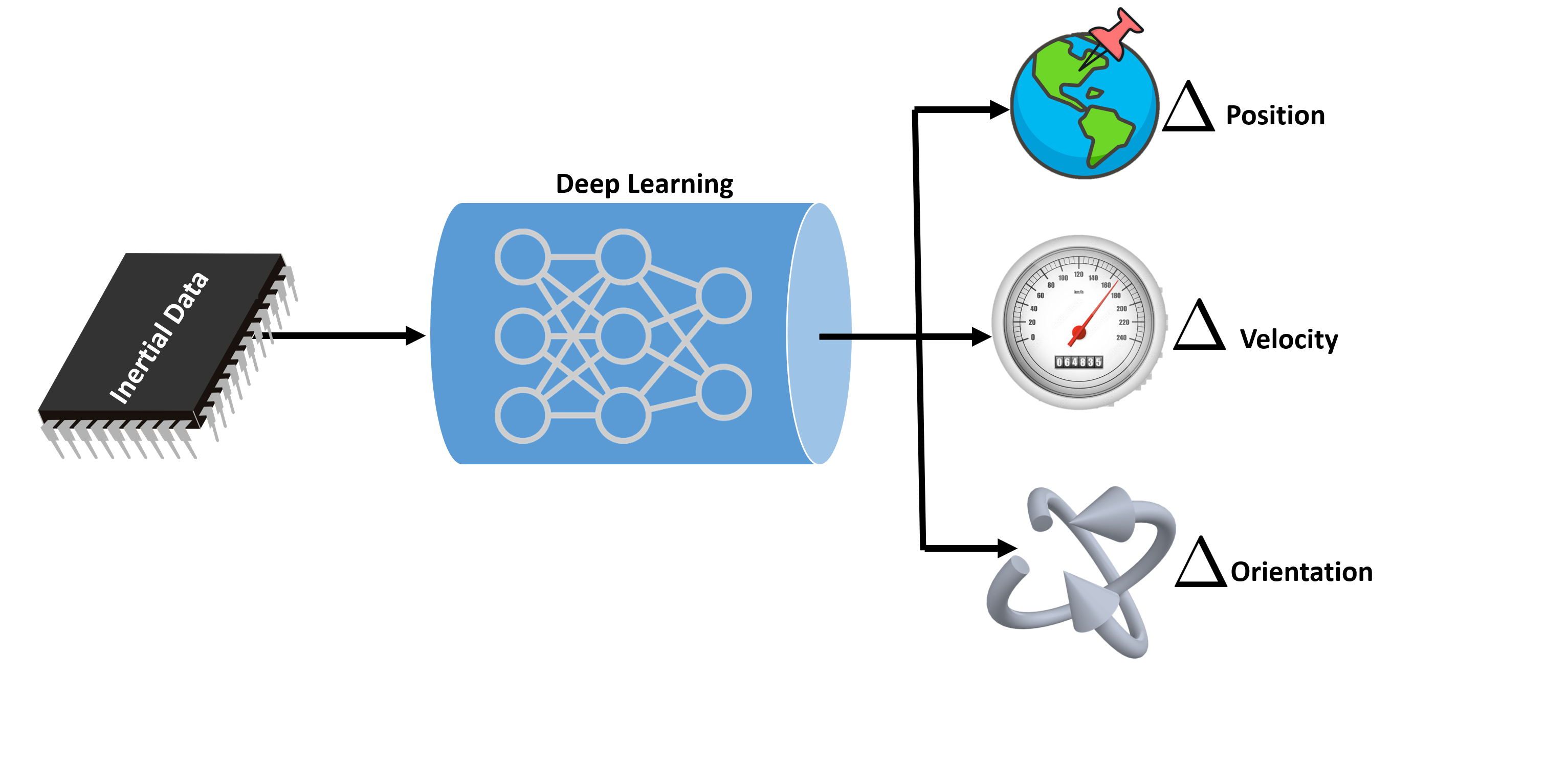} 
    \caption{Using an end-to-end DL approach to regress the residual \\ between the current state and the previous one of the vehicle.} 
    \label{fig2:b} 
    \vspace{2ex}
  \end{subfigure} 
  \begin{subfigure}[b]{0.5\linewidth}
    \centering
    \includegraphics[width=\columnwidth]{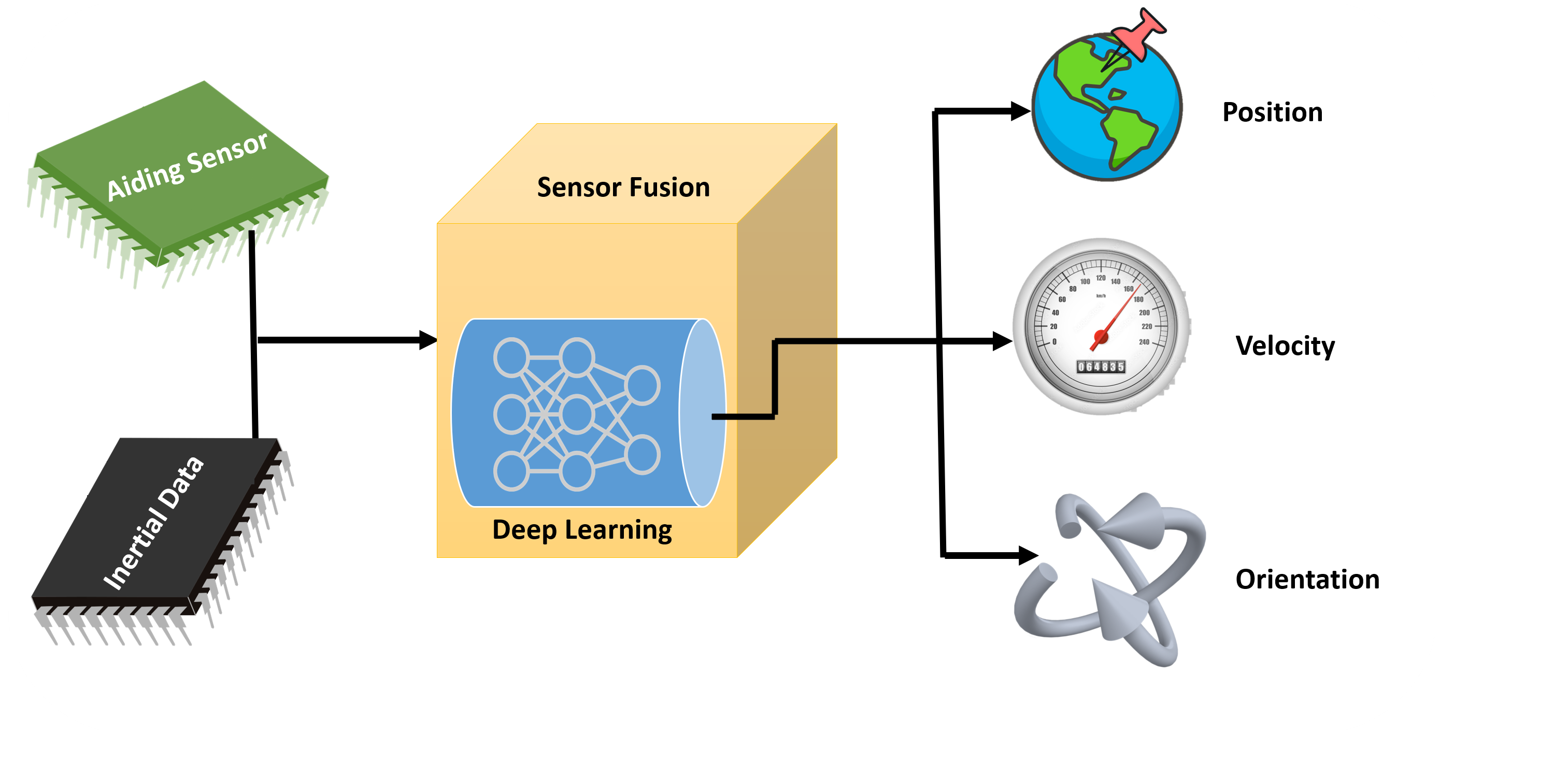} 
    \caption{Implementing sensor fusing with an end-to-end DL block to \\ regress the full state or the required increment.} 
    \label{fig2:c} 
  \end{subfigure}
  \begin{subfigure}[b]{0.5\linewidth}
    \centering
    \includegraphics[width=\columnwidth]{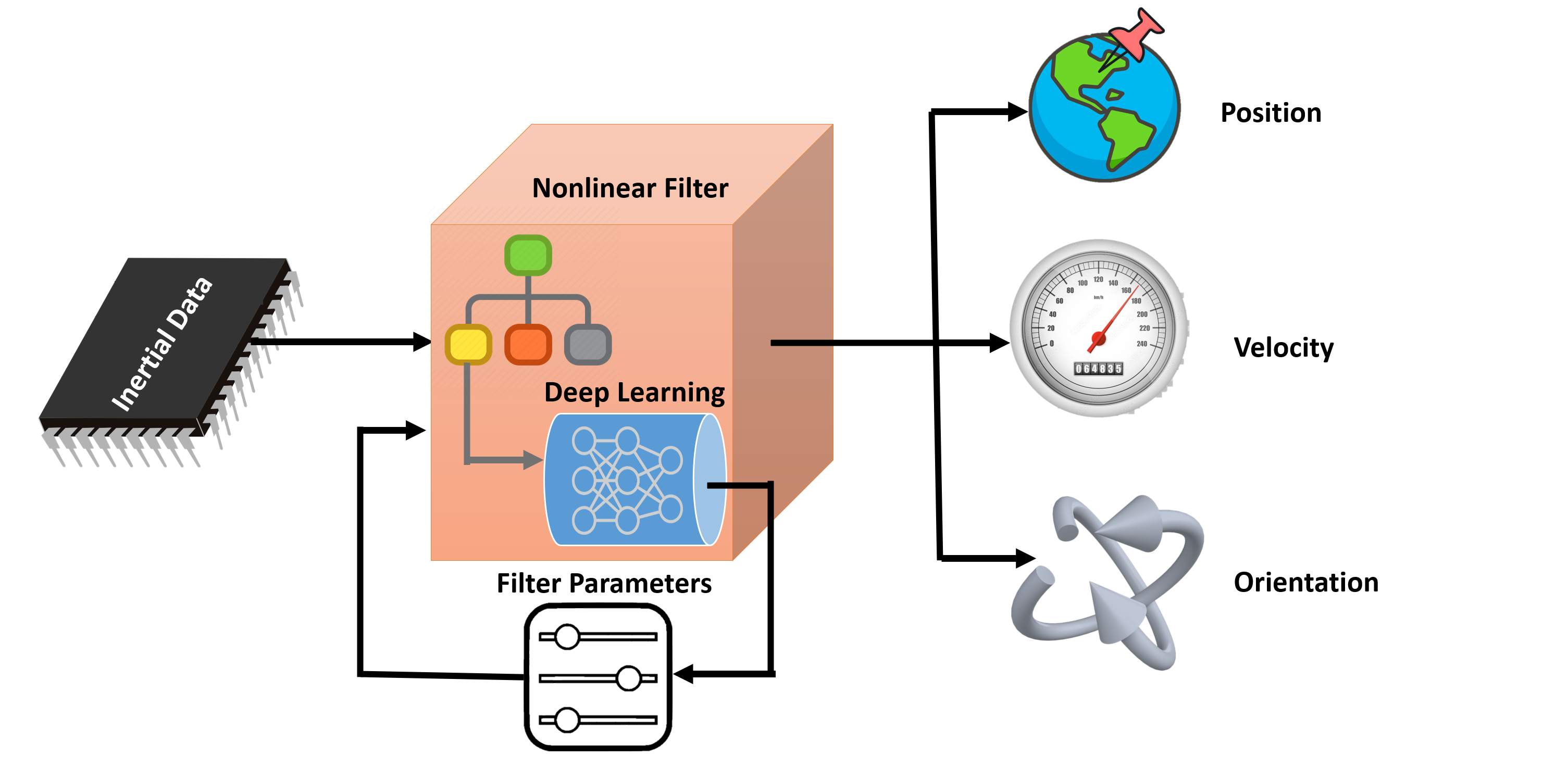} 
    \caption{In sensor fusion scenarios, DL methods are applied\\ to obtain one or more filter parameters.} 
    \label{fig2:d} 
  \end{subfigure} 
  \caption{Different techniques for improving inertial navigation using DL}
  \label{fig2} 
\end{figure*}
\noindent
In this section, we delve into the survey findings, summarizing the contributions, which encompass common techniques and approaches. Subsequently, we weigh the pros and cons of using DL approaches for inertial navigation tasks. Finally, we explore future trends in inertial navigation with deep learning..
\subsection{Summary}
\begin{figure}[h!]
	\centering
        \includegraphics[width=\columnwidth]{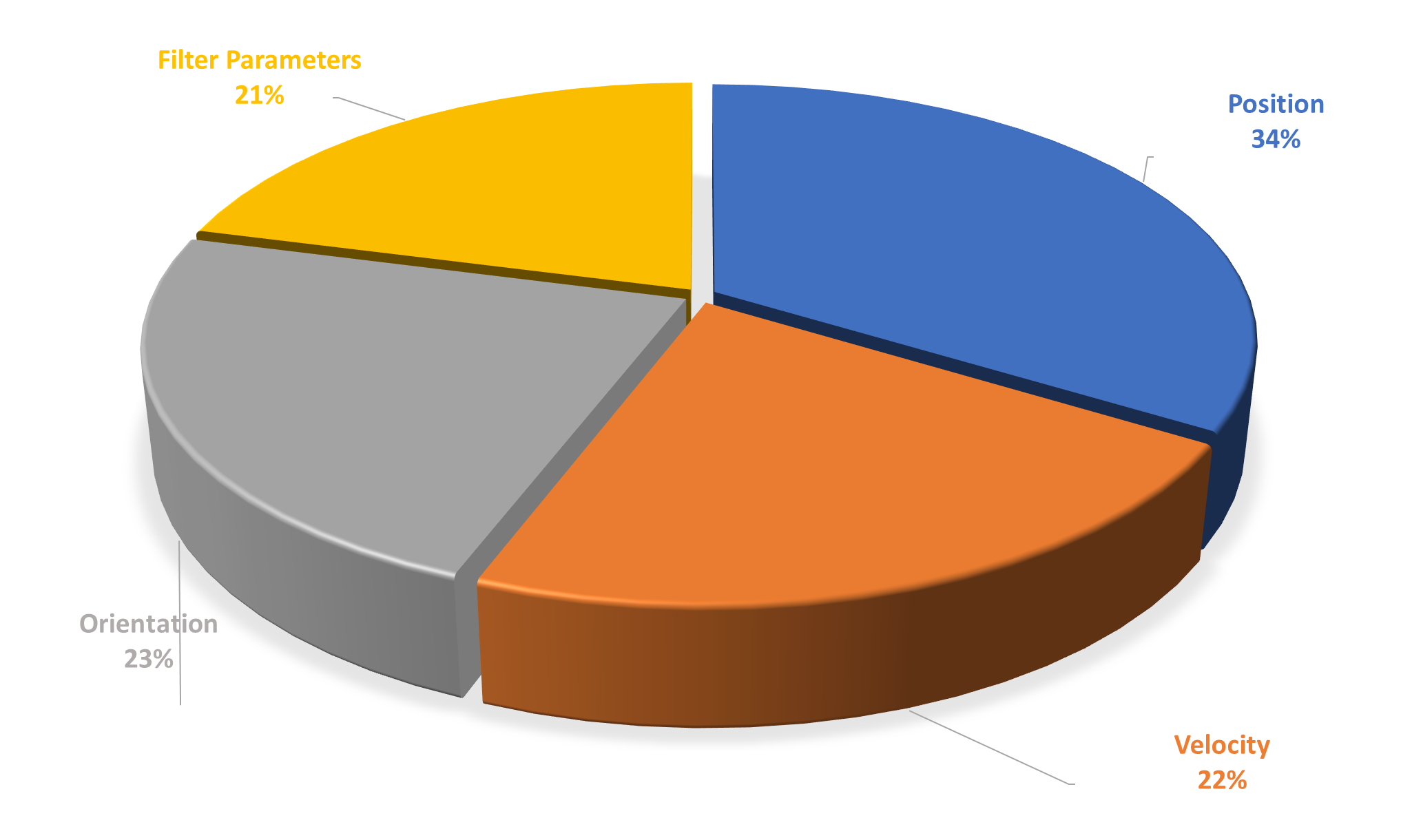}
	  \caption{DL goals in improving the navigation performance.}\label{fig3}
\end{figure}

\begin{figure}[h!]
	\centering
        \includegraphics[width=\columnwidth]{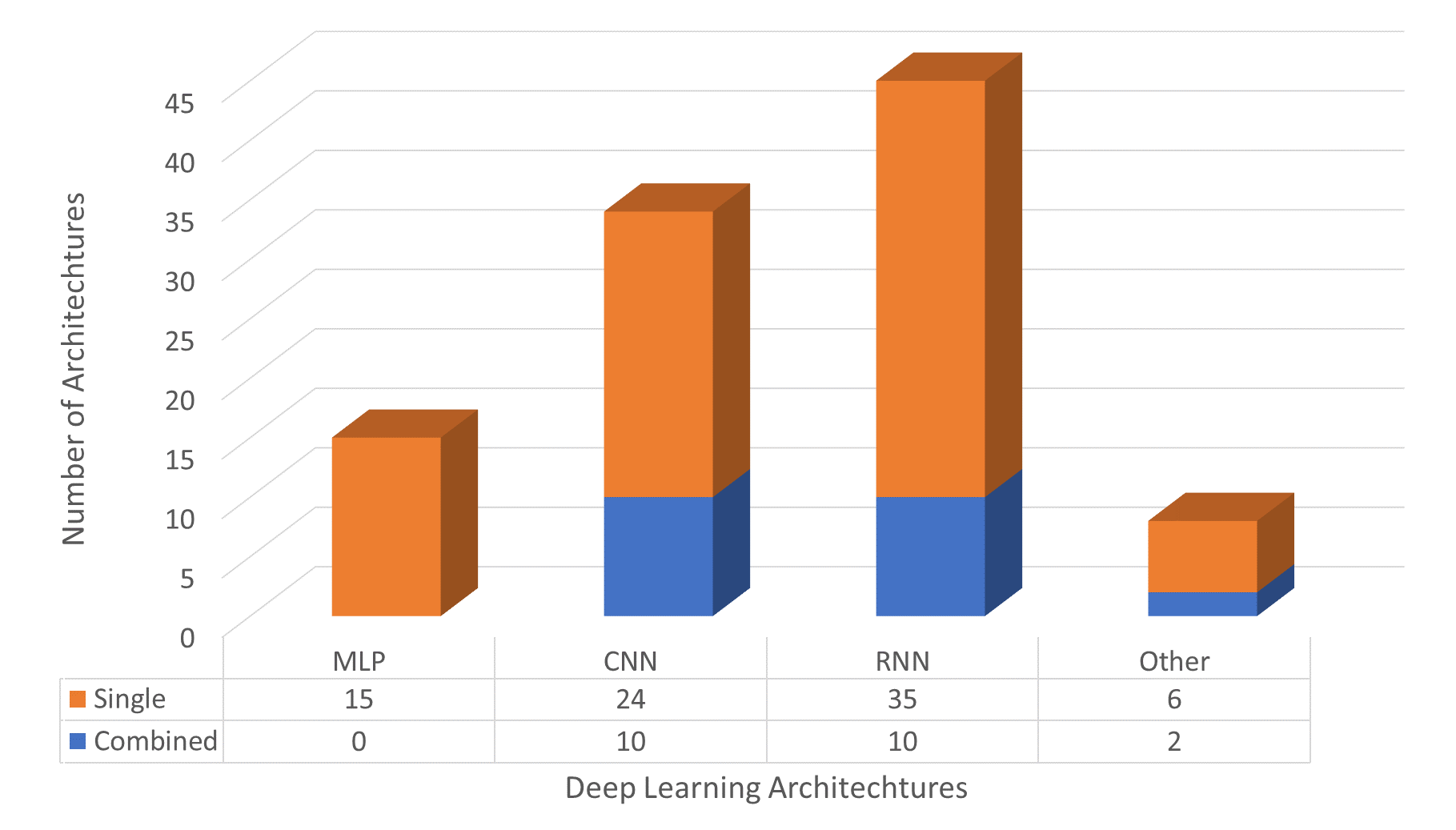}
	  \caption{Common DL architectures for improving navigation tasks, divided into single architectures and combined architectures.}\label{fig4}
\end{figure}

\noindent
The purpose of this section is to provide a comprehensive analysis of the current trends in DL methods for inertial sensing and sensor fusion, drawing insights from previously discussed studies.\\
Taking a closer look at the most common courses of action, it appears there are four repeating baseline approaches to embed inertial sensing an DL, as illustrated in Fig. \ref{fig2}.
The first approach involves inserting the inertial data into a DL architecture and regressing one or more states of the full navigation solution as shown in Fig. \ref{fig2:a}. Other studies took a different approach, focusing on analyzing the desired residual or delta required to update the current measurements, rather than regressing the entire state of navigation components. Analyzing these residuals has proven to be more effective, particularly due to the high-rate solutions provided by the inertial sensors, which often hover close to zero with a small standard deviation, resembling a normal distribution. This characteristic makes it easier for the network to handle the problem efficiently. An illustration of this concept is depicted in Fig. \ref{fig2:b}.\\
Rather than relying solely on inertial data, and as discussed before, most navigation solutions integrate inertial data with other sensors to provide a more accurate result. Fig. \ref{fig2:c} and Fig. \ref{fig2:d} show how DL is incorporated in the sensor fusion operation. The former uses both the inertial data and the aiding measurements as input to the end-to-end network to give the navigation solution. The latter target parameters of the nonlinear filter, which are responsible for the sensor fusion, such as the noise covariance matrix estimation.\\
The methods outlined in this paper are applicable across all three domains: land, aerial, and maritime, as the navigation solution remains consistent regardless of the platform or environment. While the dynamics may vary between these domains, the fundamental navigation principles remain the same. Therefore, for example, DL-based orientation estimation developed for land vehicles could be adapted for use in aerial or maritime applications with appropriate adjustments for the specific dynamics of each domain.
By examining the details of the survey, we determined what are the most common goals that current research focuses on improving, and present them in a pie chart, in Fig. \ref{fig3}. According to the chart, 79\% of the papers focus on improving position, velocity, and orientation, while 21\% addressed filter parameter improvement. Most of the latter papers were published within the past three years.
\noindent
In addition to the initial analysis, we observed the general DL architectures that have been employed. We identified four distinct architectural streams: MLP, CNN, RNN, and others. MLP includes only fully connected networks, CNN includes networks such as 1DCNN and TCN, RNN includes LSTM and GRU, and 'others' encompass architectures such as transformers, reinforcement learning, and more, which are not included in the previously mentioned categories. As shown in Fig. \Ref{fig4}, the networks are also divided into single and combined networks, where combined refers to architectures that comprise more than one method, such as CNN-RNN. The bar plot indicates that the primary architecture is RNN, along with its variations. This observation is sensible given that this architecture was explicitly developed for time-series problems, enabling it to detect temporal dependencies effectively. Furthermore, CNN methods play a significant role in inertial navigation, establishing themselves as the second most popular architecture in the field. In certain scenarios, they showcase superior accuracy when compared to RNNs. Their capability to excel at extracting informative features from small time windows, typically spanning just a few seconds, makes them particularly effective. Moreover, CNN architectures serve as the backbone for numerous denoising and calibration methods, underscoring their versatility and effectiveness within this domain. The MLP is a fundamental architecture and was one of the earliest to be adopted in inertial navigation. However, while it is common for modern networks to incorporate FC layers in the final block, the MLP itself does not excel at extracting sufficient features independently. Its simplicity may limit its effectiveness in capturing complex patterns and relationships within inertial navigation data. Despite its current status as the least popular, the 'others' category is gaining momentum, largely due to its newfound recognition. With the advancements in natural language processing, architectures such as transformers and bidirectional encoder representations from transformers (BERT) have started to surface in recent literature, displaying significant potential and delivering promising results. Notably, they have demonstrated superiority over MLPs, CNNs, and RNNs in various fields, marking a notable shift in the landscape of NN architectures.
\subsection{Pros and Cons}
\noindent
There are a number of advantages to using DL methods in inertial navigation, including: 
\begin{enumerate}
    \item \textbf{Modeling Nonlinear Problems} - Inertial navigation presents a nonlinear challenge where DL methods excel, as supported by survey findings. DL approaches have demonstrated effectiveness across various aspects of inertial navigation and have shown practical impact across different platforms and domains.
    \item \textbf{Parameter Estimation} - In nonlinear inertial sensor fusion, the survey indicates that employing DL for estimating various filter parameters, such as process or measurement noise covariance, can markedly enhance performance.
    \item \textbf{Robustness} - DL approaches exhibit the capacity to generalize across diverse scenarios, a critical attribute in navigation tasks. Variations in platform maneuvers, coupled with external factors like wind, waves, and obstacles, can be effectively learned and handled through DL techniques.
    \item \textbf{Real-time Processing} - The advancement in computational hardware, coupled with the temporal nature of inertial data, makes DL methods particularly efficient. Unlike image processing tasks, which demand heavier computational loads, inertial data processing requires fewer computational resources. Once trained, DL models can operate in real-time, offering swift and effective navigation solutions.
    \item \textbf{Multimodal Integration} - Inertial navigation often involves INS fusion with various sensors like GNSS and DVL, among others. The DL approaches highlighted in this survey illustrate how to fuse these sensors using end-to-end methods and hybrid approaches to improve the overall system performance. 
\end{enumerate}
\noindent
As well as the above benefits, there are some cons that need to be taken into account as well. Among them are:
\begin{enumerate}
    \item \textbf{Data Dependency} - DL models necessitate substantial amounts of high-quality data to effectively generalize the problem. Navigation presents a challenge as each platform exhibits unique dynamics and maneuvers. Training a model on specific dynamics may lead to overfitting, and variations in environmental conditions such as weather, wind, and temperature can influence sensor performance and dynamics, requiring adaptable learning approaches or more data.
    \item \textbf{Computational Resources} -Training DL models typically requires significant computational resources, such as high-performance GPUs or TPUs, which are often characterized by high costs and limited availability. This challenge is particularly relevant in the context of the increasing adoption of networks designed for large language models, such as transformers, in inertial sensing and sensor fusion applications.
    \item \textbf{Interpretability} - DL models are often regarded as "Black Boxes," meaning that while the inputs and outputs are known, the internal workings of the learning system remain opaque. This presents a significant challenge in safety-critical navigation applications where users rely on understanding and trusting the system's behavior.
    \item \textbf{Lack of Datasets and Benchmarks} - Although inertial navigation has a long history, publicly available inertial and reference data is lacking, often of poor quality. For example, while GNSS position references offer accuracy with a standard deviation of a few meters, GNSS-RTK data provides centimeter-level accuracy and would be more appropriate as a reference. Furthermore, the field of DL and inertial navigation lacks common benchmarks across platforms, hindering clear indications of improvement over time.
    \item \textbf{Sensor Fusion Cross-Correlation } - A common approach to using DL with inertial navigation is to enhance either the aiding updates or the actual inertial readings by fusing the inertial data with aiding sensors such as GNSS and DVL. While this approach often demonstrates good performance, it sometimes introduces cross-correlation between the sensors, which are typically ignored. Addressing these cross-correlations poses a challenge, as DL approaches inherently involve nonlinear functions. Consequently, incorporating the network's output into a nonlinear filter requires careful consideration of these cross-correlations. 
\end{enumerate}

\subsection{Future Trends}
\begin{figure}[h!]
	\centering
        \includegraphics[width=\columnwidth]{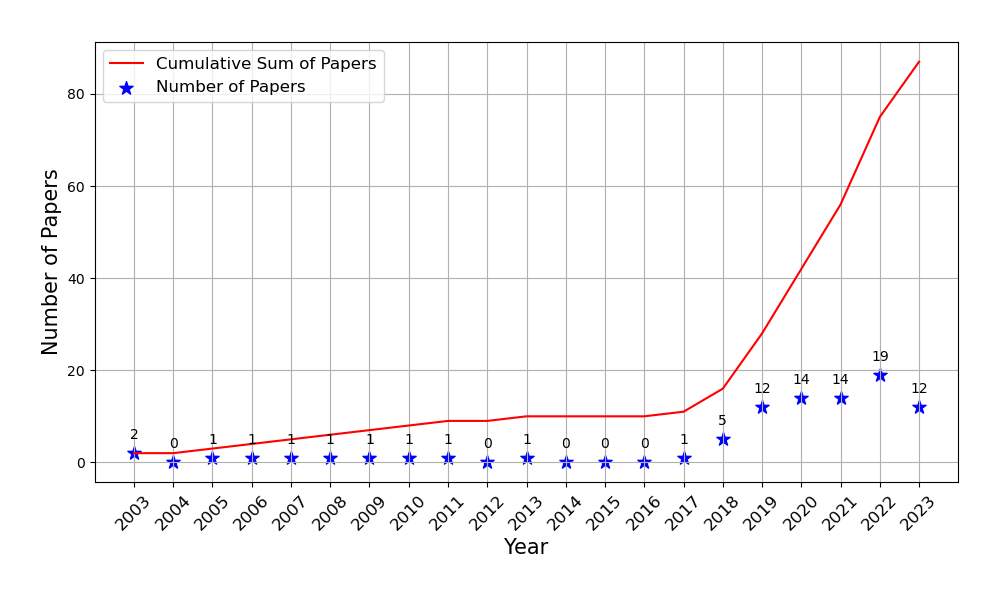}
	  \caption{The number of papers published from 2003 to 2023. Each year is represented by blue stars, while the cumulative sum of papers is depicted by the red curve.}\label{impact}
\end{figure}

\noindent
Figure \ref{impact} depicts the burgeoning trend in the field of inertial navigation aided by DL. The significant increase in publications, evident from 2019 onwards and continuing to rise, underscores the growing integration of DL approaches in this domain. During the initial phase of this trend, from 2019 to 2022, conventional architectures such as MLP, CNN, and RNN predominated. However, from 2022 onwards, there has been a noticeable shift towards employing more complex architectures, aligning with the broader trend in DL research.
\\ \noindent
Upon reviewing the papers included in the survey, a clear trajectory emerges at the intersection of DL and inertial navigation. Previously, the focus was on employing end-to-end models to directly predict navigation states such as position, velocity, and orientation. However, there has been a notable shift in recent times towards leveraging the model itself. The shift towards leveraging DL techniques in inertial navigation is evident in the adoption of DL for denoising and calibration tasks, followed by the integration of refined inertial data into the navigation filter. Additionally, there is a growing trend towards enhancing filter parameters to achieve more accurate estimates of the navigation solution. Rather than directly estimating navigation components, improvements in filter parameters, such as estimating the process or measurement noise covariance, not only enhance the navigation solution but also contribute to the DL model's deeper understanding of the underlying model. Finally, while in the past, most models relied on RNNs, CNNs, and MLPs, there is now a trend toward employing more complex models derived from large language models. Specific adaptations of these models have shown significant improvements over the previously mentioned architectures.
\section{Conclusions}\label{conc}
\noindent
Inertial navigation has garnered significant attention over the past decade due to the versatility of inertial sensors across diverse platforms and environments. Historically, navigation solutions have relied on model-based algorithms. However, there has been a notable shift towards data-driven methods, particularly with the increasing popularity and capabilities of DL techniques. This integration of DL represents a significant advancement in the approach to developing navigation solutions.
This paper provides a comprehensive survey of DL methods applied to inertial navigation, specifically focusing on different platforms and practical applications. It reviewed research conducted across three distinct domains: land, aerial, and maritime. Additionally, the paper delves into calibration and denoising methods within the context of inertial navigation and DL. Furthermore, it offers insights into the trajectory of research in this area through statistical analysis.\\ \noindent
Our findings indicate that the majority of research in this area has focused on land vehicles rather than aerial or maritime vehicles, or on calibration and denoising techniques. However, some papers suggest that despite differences in mechanics, maneuvers, etc., techniques can be adapted to various platforms, as the task of navigation remains consistent across all platforms, and data-driven networks can potentially learn these differences. While most reviewed papers aimed to enhance one or more aspects of the inertial navigation algorithm for improved solutions, recent years have seen a shift towards improving filter parameters for enhanced sensor fusion processes and increased reliance on the algorithm, incorporating mathematical models as well. Although leading DL architectures have traditionally been based on RNNs and CNNs, recent research has been inspired by approaches from natural language processing, importing and adapting leading architectures from that field.
\\ \noindent
In conclusion, since 2019, there has been a notable surge in the utilization of DL methods for inertial navigation applications. These approaches have demonstrated superior performance compared to traditional model-based techniques, indicating significant potential for future research in inertial sensing. This evolution suggests a promising trajectory for further advancements in the field of inertial navigation aided by DL algorithms.

\section*{Acknowledgment}
\noindent
N.C. is supported by the Maurice Hatter Foundation and
University of Haifa presidential scholarship for outstanding students on a direct Ph.D. track.

\bibliographystyle{IEEEtran}
\bibliography{bio.bib}

\end{document}